\documentclass[11pt]{article}

\usepackage[final]{acl}

\usepackage{times}
\usepackage{latexsym}
\usepackage{multirow}
\usepackage{booktabs} 
\usepackage{hyperref}

\usepackage[T1]{fontenc}

\usepackage[utf8]{inputenc}

\usepackage{microtype}
\usepackage{textalpha}

\usepackage{inconsolata}

\usepackage{graphicx}

\title{Can I guess where you are from? \\ Modeling dialectal morphosyntactic similarities in Brazilian Portuguese}

\author{
Manoel Siqueira \\
Universidade Federal de Alagoas \\
Delmiro Gouveia - AL - Brazil \\
\texttt{manoelsqr@gmail.com}
\And
Raquel Freitag \\
Universidade Federal de Sergipe \\
São Cristóvão - SE - Brazil \\
\texttt{rkofreitag@academico.ufs.br}
}

\begin{document}
\maketitle
\begin{abstract}
This paper investigates morphosyntactic covariation in Brazilian Portuguese (BP) to assess whether dialectal origin can be inferred from the combined behavior of linguistic variables. Focusing on four grammatical phenomena related to pronouns, correlation and clustering methods are applied to model covariation and dialectal distribution. The results indicate that correlation captures only limited pairwise associations, whereas clustering reveals speaker groupings that reflect regional dialectal patterns. Despite the methodological constraints imposed by differences in sample size requirements between sociolinguistics and computational approaches, the study highlights the importance of interdisciplinary research. Developing fair and inclusive language technologies that respect dialectal diversity outweighs the challenges of integrating these fields.

\end{abstract}

\section{Introduction}

The ability to discriminate dialectal varieties of a language is an important skill for both individuals and natural language processing systems. Knowing where a person is from by the way they speak has origins so remote that even in the Bible this ability is explored in the shibboleth episode (Judges 12:5-6), and in the field of language studies, quantitative sociolinguistics has been dedicated to identifying correlations between linguistic cues and social profiles since the pioneering work of Labov (\citeyear{ labov1972sociolinguistic, labov2006social}).

The identification of linguistic features that reveal characteristics of who produced an utterance (age, gender, geographic location, among others) can be applied in the configuration and personalization processes of different artificial intelligence tasks. Knowledge of how dialectal varieties are identified in society by individuals is important for AI applications; at the same time, machine learning techniques can contribute to improving linguistic descriptions to identify patterns of lectal coherence \cite{guy2013cognitive, oushiro2016social, vaughn2019stylistically, thukroo2022review}.

The identification of dialectal profiles has implications that go beyond linguistic description, touching on ethical issues in the development of language technologies. Speakers of different dialectal varieties often face negative stereotypes. Recent studies show that large language models mirror these biases, exhibiting negative associations with dialects and amplifying discrimination when dialectal features are explicitly mentioned \cite{fleisig2024linguistic, bui2025large, lin2025assessing, abboud2024towards}. These biases are also present in Brazilian Portuguese \cite{freitag2024performance}. 

Thus, this paper investigates, from a sociolinguistic perspective, how variable morphosyntactic linguistic features co-occur within speech communities, comparing statistical approaches of correlation and clustering to understand the relationships among four phenomena in actual language use, using the \textit{Displacements} samples \cite{siqueira_freitag_2025} from \textit{Falares Sergipanos} \cite{freitag2013falares}, a sociolinguistic dataset; and tests, from an NLP perspective, whether models based on morphosyntactic embeddings can group Brazilian Portuguese speakers according to regional patterns.

\section{Sociolinguistic background}

Research on linguistic variation has traditionally focused on the analysis of single variables, aiming to identify structural or social constraints that act on a particular feature. Variationist studies have traditionally examined linguistic variables independently to establish correlations with social and structural factors such as age, gender, and regional origin \cite{labov1972sociolinguistic, labov2006social}. In Brazilian Portuguese, these studies have documented extensive variability in pronoun and agreement patterns \cite{scherre2015, oushiro2016social, scherre2021}.

However, recent developments in sociolinguistics demonstrate that dialectal organization cannot be fully understood when linguistic variables are examined in isolation \cite{guy2013cognitive}. Instead, variables often co-occur systematically, forming patterns of covariation that reflect underlying regularities within speech communities. This shift in perspective emphasizes that speakers do not index identity through a single linguistic variable, but through a pattern of linguistic variables used jointly in discourse. From the perspective of speaker behavior, identification is not achieved through the occurrence of a single feature, but rather through a constellation of linguistic traits. Covariation captures how variables interact and form coherent patterns that contribute to dialectal identity.

\begin{figure*}[h]
    \centering
    \includegraphics[width=0.8\textwidth]{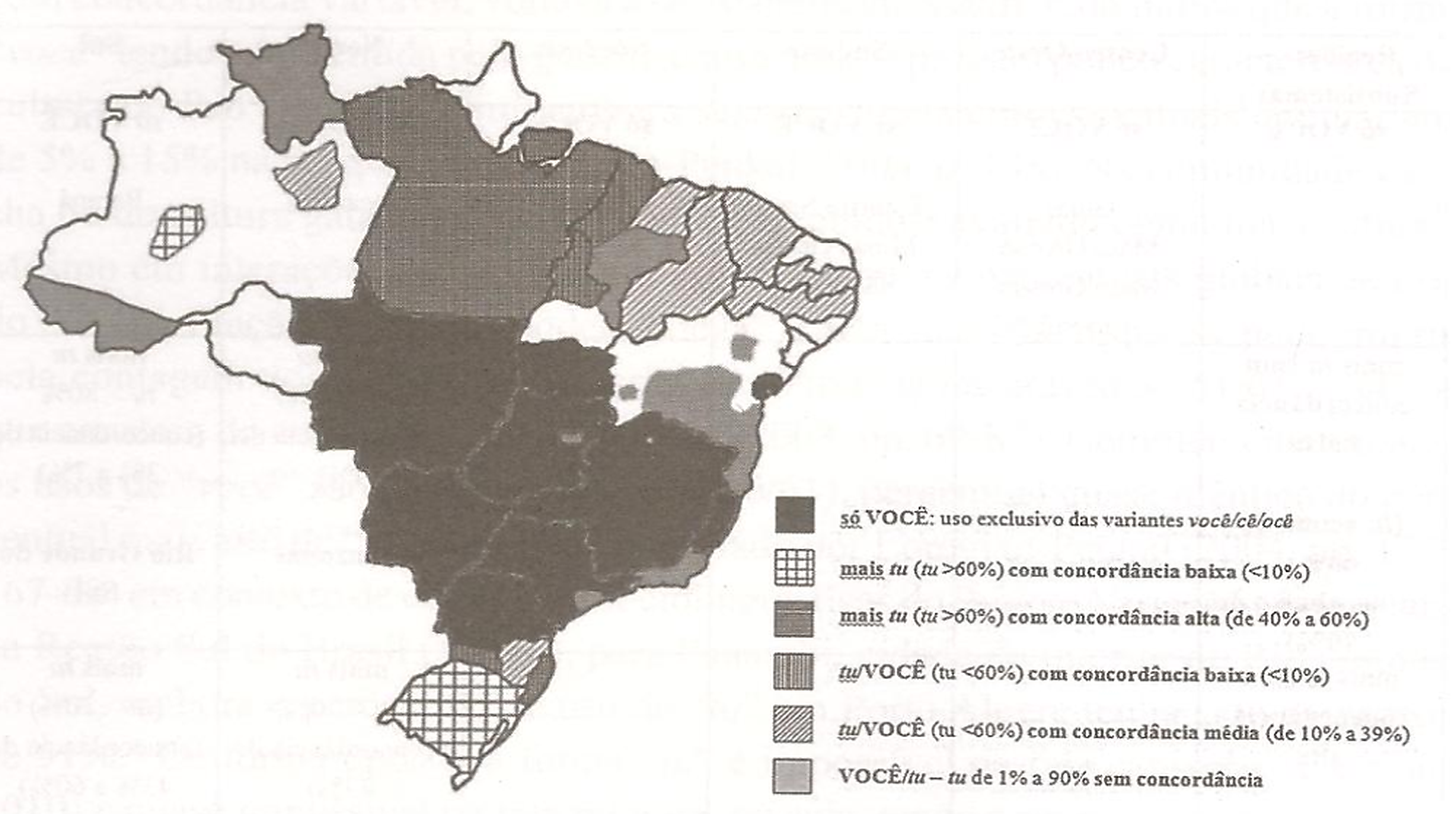}
    \caption{Scherre et al. dialectal distribution of 2P in Brazil, \cite{scherre2015}}
    \label{fig:Fig1}
\end{figure*}

In Brazilian Portuguese morphosyntactic variables related to second-person pronouns (2P) are particularly illustrative of this phenomenon. The distribution of second-person pronouns does not operate independently: the presence of one form often predicts the use or avoidance of another \cite{scherre2015, scherre2021}. Scherre et al. (\citeyear{scherre2021}) present a distribution of frequency patterns for two structurally correlated variables based on indirect results from other studies, demonstrating that identifying a dialectal profile requires seeking structurally correlated variables. Previous research has approached this phenomenon either through indirect results from multiple studies, as in Scherre et al. (\citeyear{scherre2015}), or by focusing on specific variables individually (Figure~\ref{fig:Fig1}), such as Oushiro (\citeyear{oushiro2016social}) analysis of agreement variation.

However, dialectal profiling encompasses much more than pronoun choice and agreement patterns alone. Advancing this field, the present study proposes to identify the dialectal origin of speakers by integrating the analysis of four morphosyntactic phenomena: the second-person personal pronoun in subject position (\textbf{pro2P}); the second-person clitic pronoun (\textbf{clit2P}); the second-person possessive pronoun (\textbf{poss2P}); and the presence/absence of a determiner before a possessive (\textbf{det-poss}) (Table~\ref{tab:2ps-variables}).

\begin{table}[htbp]
\centering
\begin{tabular}{ll}
\toprule
\textbf{Variable} & \textbf{Variants} \\
\midrule
\multirow{3}{*}{pro2P} & você \\
                        & cê \\
                        & tu \\
\midrule
\multirow{2}{*}{clit2P} & se \\
                         & lhe \\
\midrule
\multirow{4}{*}{poss2P} & seu (masc.) \\
                         & sua (fem.) \\
                         & teu (masc.) \\
                         & tua (fem.) \\
\midrule
\multirow{2}{*}{det-poss} & Ø seu/sua/teu/tua \\
                           & \textsc{art} seu/teu; a sua/tua \\
\bottomrule
\end{tabular}
\vspace{0.3em}
\caption{Second-person singular morphosyntactic variables}
\label{tab:2ps-variables}
\end{table}

These four variables and their levels result in 24 possible combinations (Table~\ref{tab:combinations}). Studies have shown that this combination of features interacts as a clue of lectal coherence \cite{freitag2022mobility, siqueira2023, silva_2025_covariacao}, suggesting distinct dialectal profiles across Brazilian Portuguese varieties. 

\begin{table}[h]
\centering
\small
\setlength{\tabcolsep}{3pt}
\begin{tabular}{@{}llll@{}}
\hline
\textbf{pro2P} & \textbf{clit2P} & \textbf{det-poss} & \textbf{poss2P} \\ \hline
\textit{Você} disse que & \textit{te} comprometeu com & o \textsc{art} & \textit{teu} projeto \\
\textit{Você} disse que & \textit{te} comprometeu com & Ø & \textit{teu} projeto \\
\textit{Você} disse que & \textit{te} comprometeu com & o \textsc{art} & \textit{seu} projeto \\
\textit{Você} disse que & \textit{te} comprometeu com & Ø & \textit{seu} projeto \\
\textit{Você} disse que & \textit{se} comprometeu com & o \textsc{art} & \textit{teu} projeto \\
\textit{Você} disse que & \textit{se} comprometeu com & Ø & \textit{teu} projeto \\
\textit{Você} disse que & \textit{se} comprometeu com & o \textsc{art} & \textit{seu} projeto \\
\textit{Você} disse que & \textit{se} comprometeu com & Ø & \textit{seu} projeto \\

\textit{Cê} disse que & \textit{te} comprometeu com & o \textsc{art} & \textit{teu} projeto \\
\textit{Cê} disse que & \textit{te} comprometeu com & Ø & \textit{teu} projeto \\
\textit{Cê} disse que & \textit{te} comprometeu com & o \textsc{art} & \textit{seu} projeto \\
\textit{Cê} disse que & \textit{te} comprometeu com & Ø & \textit{seu} projeto \\
\textit{Cê} disse que & \textit{se} comprometeu com & o \textsc{art} & \textit{teu} projeto \\
\textit{Cê} disse que & \textit{se} comprometeu com & Ø & \textit{teu} projeto \\
\textit{Cê} disse que & \textit{se} comprometeu com & o \textsc{art} & \textit{seu} projeto \\
\textit{Cê} disse que & \textit{se} comprometeu com & Ø & \textit{seu} projeto \\

\textit{Tu} disse que & \textit{te} comprometeu com & o \textsc{art} & \textit{teu} projeto \\
\textit{Tu} disse que & \textit{te} comprometeu com & Ø & \textit{teu} projeto \\
\textit{Tu} disse que & \textit{te} comprometeu com & o \textsc{art} & \textit{seu} projeto \\
\textit{Tu} disse que & \textit{te} comprometeu com & Ø & \textit{seu} projeto \\
\textit{Tu} disse que & \textit{se} comprometeu com & o \textsc{art} & \textit{teu} projeto \\
\textit{Tu} disse que & \textit{se} comprometeu com & Ø & \textit{teu} projeto \\
\textit{Tu} disse que & \textit{se} comprometeu com & o \textsc{art} & \textit{seu} projeto \\
\textit{Tu} disse que & \textit{se} comprometeu com & Ø & \textit{seu} projeto \\
\hline
\end{tabular}
\caption{Linear combinations of second-person subject, clitic, determiner and possessive in Brazilian Portuguese}
\label{tab:combinations}
\end{table}

This sociolinguistic background provides the theoretical and methodological foundation for the NLP methods adopted for the approach.

\section{Method}

\subsection{Data}

To observe distinct behavioral patterns of speaker groups from different regions, it is necessary to use sociolinguistic samples that allow for this type of visualization. Thus, the selection and/or stratification of the sample must consider individuals from multiple geographic areas based on research interests. One problem in sociolinguistic research is the limitation of samples: they generally follow the interview model, lasting about 1 hour of speech, which generates a large volume of linguistic data, but from a single speaker. For classification analyzes, the large volume of data from each speaker is necessary; but for clustering, a large number of speakers represented in the sample is also needed.

To test how the co-occurrence pattern of the 4 variables can provide clues to dialectal origin, this study employed semi-spontaneous speech data from three samples from the \textit{Falares Sergipanos} dataset: \textit{Deslocamentos 2019}, \textit{Deslocamentos 2020}, and \textit{Linguagem Corporificada 2023}, with 181 speakers and 1,002,535 words.

The samples consider the speech of university students from the Federal University of Sergipe (UFS), aged between 18 and 30 years (\textit{M} = 21 years). The samples were selected because they include speakers from different regions and take into account students' access to campus in terms of mobility, which allows for the indirect observation of dialectal meaning of variation based on differences in usage among speakers who belong to distinct groups, as well as the description of different dialects of Brazilian Portuguese.

\begin{table*}
  \centering
\small
\setlength{\tabcolsep}{3pt}
\begin{tabular}{|l|p{10cm}|}
    \hline
\textbf{Displacement Type} & \textbf{Description} \\
\hline
\textit{Displacement 1} & UFS students born in Aracaju's Metropolitan region (Aracaju, Nossa Senhora do Socorro, São Cristóvão, and Barra dos Coqueiros cities) and who reside there \\
\hline
\textit{Displacement 2} & UFS students born in the countryside of Sergipe who make the daily commute to UFS \\
\hline
\textit{Displacement 3} & UFS students born in the countryside of Sergipe who reside in Aracaju's Metropolitan region \\
\hline
\textit{Displacement 4} & UFS students born in other states (predominantly Bahia and Alagoas) who currently reside in Aracaju's Metropolitan region \\
\hline
\end{tabular}
\caption{Control of displacements and dialectal boundaries}
\label{tab:displacements}
\end{table*}

\begin{table*}[h]
\centering
\small
\begin{tabular}{|p{1.2cm}|p{9cm}|p{4.8cm}|}
\hline
\textbf{Variable} & \textbf{Rules} & \textbf{Context} \\
\hline
det-poss & 
\texttt{[\{'LEMMA': \{'NOT\_IN': ['meu','teu', 'seu', 'nosso']\}\}, \{'POS': 'DET', 'MORPH':\{'IS\_SUPERSET': ['PronType=Prs']\}\}, \{'POS': 'NOUN'\}]} & 
Possessives \textit{meu}, \textit{teu}, \textit{seu}, and \textit{nosso} and inflections) in contexts where they precede nouns. \\
\hline
pro2P & 
\texttt{[\{'ORTH': \{'IN': ['você', 'cê', 'tu']\}\}, \{'POS': 'VERB', 'MORPH': \{'IS\_SUPERSET': ['VerbForm=Fin']\}\}]}

\texttt{[\{'ORTH': \{'IN': ['você', 'cê', 'tu']\}\}, \{\}, \{'POS': 'VERB', 'MORPH': \{'IS\_SUPERSET': ['VerbForm=Fin']\}\}]}
& 
Pronouns \textit{você}, \textit{tu}, and \textit{cê} preceding verbs.

Pronouns \textit{você}, \textit{tu}, and \textit{cê} preceding verbs with intervening material between pronoun and verb. \\
\hline

clit2P & 
\texttt{[\{'ORTH': \{'IN': ['te', 'lhe']\}\}, \{'POS': 'VERB'\}]}

\texttt{[\{'POS': 'VERB'\}, \{'ORTH': \{'IN': ['te', 'lhe']\}\}]}

\texttt{[\{'POS': 'VERB'\}, \{'IS\_PUNCT': True\}, \{'ORTH': \{'IN': ['te', 'lhe']\}\}]} & 

Pronouns \textit{te} and \textit{lhe} preceding verbs in proclitic position.

Pronouns \textit{te} and \textit{lhe} following verbs in enclitic position.

Pronouns \textit{te} and \textit{lhe} following verbs in enclitic position, connected by a punctuation mark (-). \\
\hline

poss2P & 
\texttt{[\{"LEMMA": \{"IN": ["seu", "teu"]\}\}, \{"POS": "NOUN"\}]}

\texttt{[\{"POS": "NOUN"\}, \{"LEMMA": \{"IN": ["seu", "teu"]\}\}]} & 

Pronouns \textit{teu} and \textit{seu} (and inflections) preceding nouns.

Pronouns \textit{teu} and \textit{seu} (and inflections) following nouns. \\

\hline
\end{tabular}
\caption{Search rules in dataset}
\label{tab:search_rules}
\end{table*}

The collection of samples follows the protocol for sociolinguistic interviews that last approximately 40 to 60 minutes and is based on a script of diverse questions. The first questions are used to verify factual information about the speaker, while the others address social issues such as education, security, health, and gender equality.

The samples were constituted following a stratified random sampling method, in which we randomly recruited students from the campus, and also convenience sampling, given the collaboration of volunteer students and their accessibility \cite{buchstaller2013population, freitag2018amostras}. All interviews were transcribed by humans with the support of ELAN software \cite{hellwig2003eudico}. After this process, all transcriptions were exported in the program's format and in .txt.

The occurrences of each variable were extracted with spaCy library, and exported using the data.frame function from the pandas library to store the data in a spreadsheet that, in the future, can be used on other platforms, such as R, for descriptive and inferential statistical analyses. The spreadsheet columns present the following information: order of occurrence in the files, name of the *.txt file from which the searched term was extracted, preceding and following context, and the complete context where the searched term is found.

After automatic extraction and classification, data validation was also performed manually by human annotators. Once the spreadsheets with occurrences of the morphosyntactic phenomena were organized, statistical analysis was conducted. 

\subsection{Statistical analysis}

Statistical analyses were conducted using R (R Core Team, 2025). The significance level was set at α < 0.05, following standard practices in social, human, and cognitive sciences. Application values for each variable were established based on predominant forms identified in previous studies conducted in Sergipe \cite{siqueira2023}: \textbf{absence of determiner} before possessives (\textit{det-poss}), \textbf{você} for subject pronouns (\textit{pro2PS}), \textbf{te} for clitics (\textit{cli2PS}), and \textbf{seu} for possessives (\textit{pos2PS}).

The analytical approach comprised three main stages: classification through univariate analysis examining each phenomenon individually, correlation analysis investigating relationships between variables, and clusterization identifying speaker groups based on joint patterns across multiple variables.

\textbf{Classification}: Chi-square tests (or Fisher's exact test when conditions were not met) assessed relationships between each morphosyntactic variable and social factors (mobility, time in program, gender, and age). Cramer's V² measured association strength when significant relationships were found, with values ranging from 0 (no association) to 1 (strong association). Logistic regression modeled the relationship between age and categorical variables. Variables were operationalized as continuous numeric values extracted in two ways: (i) usage rates of application values calculated per individual speaker, and (ii) log odds extracted from mixed-effects models with speaker as random effect using the lme4 package \cite{bates2015fitting}.

\textbf{Correlation}: This stage employed two methodological approaches. First, Spearman's \textit{rho} correlation coefficients examined whether linguistic phenomena correlate with each other, following \citeauthor{guy2013cognitive} and \citeauthor{oushiro2016social}. Data distribution was assessed using the Shapiro-Wilk test and histogram visualization. Second, social clustering analysis classified individual frequency rates into ternary categories (Low $<$ 40\%, Medium 40--60\%, High $>$ 60\%) or binary categories (High $\geq$ 50\%, Low $<$ 50\%).

\textbf{Clusterization}: K-medoids cluster analysis identified speaker groups based on joint usage patterns of the four morphosyntactic variables, following \citet{freitag2022mobility}. Clustering was performed using the pam function (partitioning around medoids) from the cluster package \cite{maechler2023} to group the dataset obtained through individual mean usage rates of the four variables.

Dataset and code are available at: 
\href{https://osf.io/6nk4q/overview?view_only=1db7d39504084bf4a7ec08cb0ab65d00}{https://osf.io/6nk4q/}

\section{Results}

\subsection{Classification}

The dataset comprises 181 sociolinguistic interviews, with speakers classified by gender, dialectal origin, time-in-program and age. 
The classification analysis examined four morphosyntactic variables related to second-person pronouns in Brazilian Portuguese (Table~\ref{tab:classification}). 

For \textbf{pro2P}, \textit{você} was the predominant form with 79.6\% (3064/3834 tokens), followed by the reduced variant \textit{cê} with 21.5\% (770/3848), while \textit{tu} showed marginal usage at 0.4\% (14/3848).

Regarding \textbf{clit2P}, \textit{te} occurs with higher frequency at 71.9\% (194/270), whereas \textit{se} accounted for 28.1\% (76/270). 
Of all speakers, only 12 show variable use of \textit{te} and \textit{lhe}; categorical use of a single form is the rule, particularly \textit{te}.

For possessive pronouns (\textbf{poss2P}), \textit{seu} is the most frequent possessive pronoun at 99.1\% (528/533), with categorical use being the norm in our data, while \textit{teu} represents only 0.9\% (5/533).

For \textbf{det-poss}, determiner absence occurred in 53.2\% of cases (4313/8114), while the presence of a determiner (\textsc{art}) was observed in 46.8\% (3801/8114). 
Individual speaker rates were calculated for this variable.

To observe possible dialectal distinctions and salience effects, mobility and time-in-program variables were employed. 
Dialectal distinctions are visible in three of the four phenomena in at least one sample: variable use of determiners before possessives \textbf{det-poss}, variation in 2PS subject pronouns (excluding \textit{tu}) \textbf{pro2P}, and variation in 2P clitics \textbf{clit2P}, all of which show distinct usage patterns across dialectal groups.

Regarding dialectal strength, changes in speakers’ linguistic behavior toward the end of their program indicate recognition — albeit not necessarily conscious — of different uses of morphosyntactic variables. 
This provides indirect evidence of salience through exposure to and contact with divergent usage patterns.

Analysis of macro-social variables such as gender and age predominantly indicated non-significant effects on the distribution of the morphosyntactic phenomena examined.

\begin{table}[!t]
\centering
\begin{tabular}{llcc}
\toprule
\textbf{Variable} & \textbf{Variant} & \textbf{Frequency} & \textbf{\%} \\
\midrule
pro2PS  & \textit{você} & 3064/3834 & 79.6 \\
        & \textit{cê}   & 770/3848  & 21.5 \\
        & \textit{tu}   & 14/3848   & 0.4  \\
\midrule
clit2PS & \textit{te} & 194/270 & 71.9 \\
        & \textit{se} & 76/270  & 28.1 \\
\midrule
poss2PS & \textit{seu} & 528/533 & 99.1 \\
        & \textit{teu} & 5/533   & 0.9  \\
\midrule
det-poss & Ø   & 4313/8114 & 53.2 \\
         & \textsc{art} & 3801/8114 & 46.8 \\
\bottomrule
\end{tabular}

\vspace{0.3em}
\footnotesize
\textit{Note.} Dataset comprises 181 sociolinguistic interviews classified by gender, dialectal origin, and age.

\caption{Distribution of second-person singular morphosyntactic variants}
\label{tab:classification}
\end{table}

\begin{figure*}[h]
    \centering
    \includegraphics[width=0.8\textwidth]{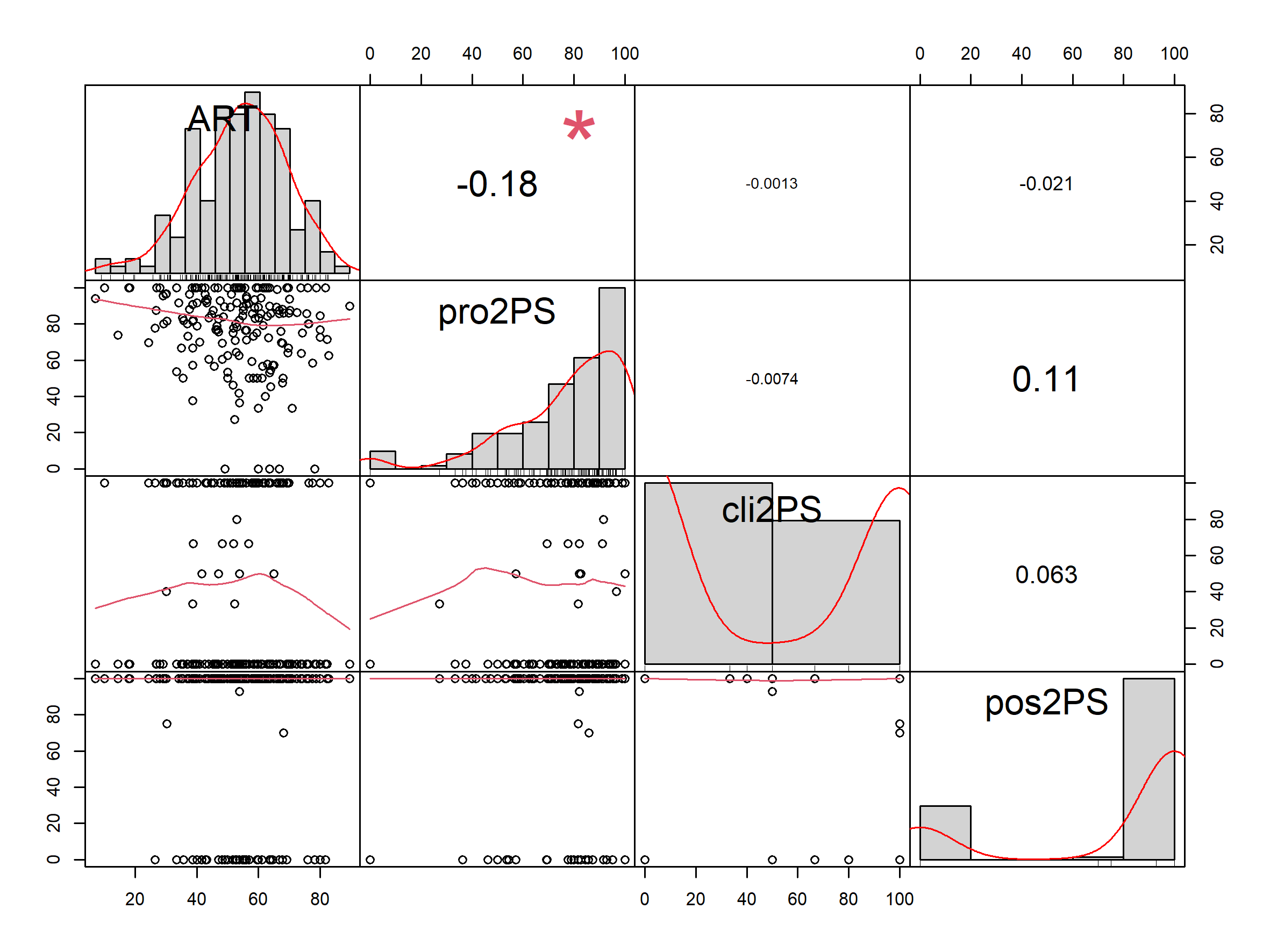}
    \caption{Correlation results}
    \label{fig:Fig2}
\end{figure*}

\subsection{Correlation}

Correlation analysis examined the four morphosyntactic variables through two methods. First, individual usage frequencies of the application variant were calculated for each speaker automatically. Speakers without occurrences received a 0\% rate. Only determiner absence showed normal distribution according to the Shapiro-Wilk test (\(p = 0.21\)). 

Second, log odds were extracted through mixed-effects models using the \texttt{glmer} function from the \texttt{lme4} package, with speaker as a random effect. Only determiner absence (\(p = 0.11\)) and \textit{você} (\(p = 0.34\)) showed normal distribution.

Due to violation of the normality assumption in most data, the non-parametric Spearman correlation test was employed. The rho coefficient (\(\rho\)) ranges from -1 to +1, where values close to the extremes indicate strong correlation, values near 0 indicate weak or no correlation, positive signs indicate direct relationship, and negative signs indicate inverse relationship.

Results revealed significant negative correlation only between determiner absence (\textbf{det-poss}) and the pronoun \textit{você} (\textbf{pro2PS}), both in frequencies (\(\rho = -0.18, p = 0.017\)) and log odds (\(\rho = -0.16, p = 0.036\)) (Figure~\ref{fig:Fig2}). This indicates that speakers who frequently use determiner absence tend not to use \textit{você} as 2PS subject, possibly reflecting specific linguistic choices from their geographic region. 

For the remaining variable pairs, including pronoun variables expected linguistically, no significant correlation was found.

\begin{figure*}[h]
    \centering
    \includegraphics[width=0.8\textwidth]{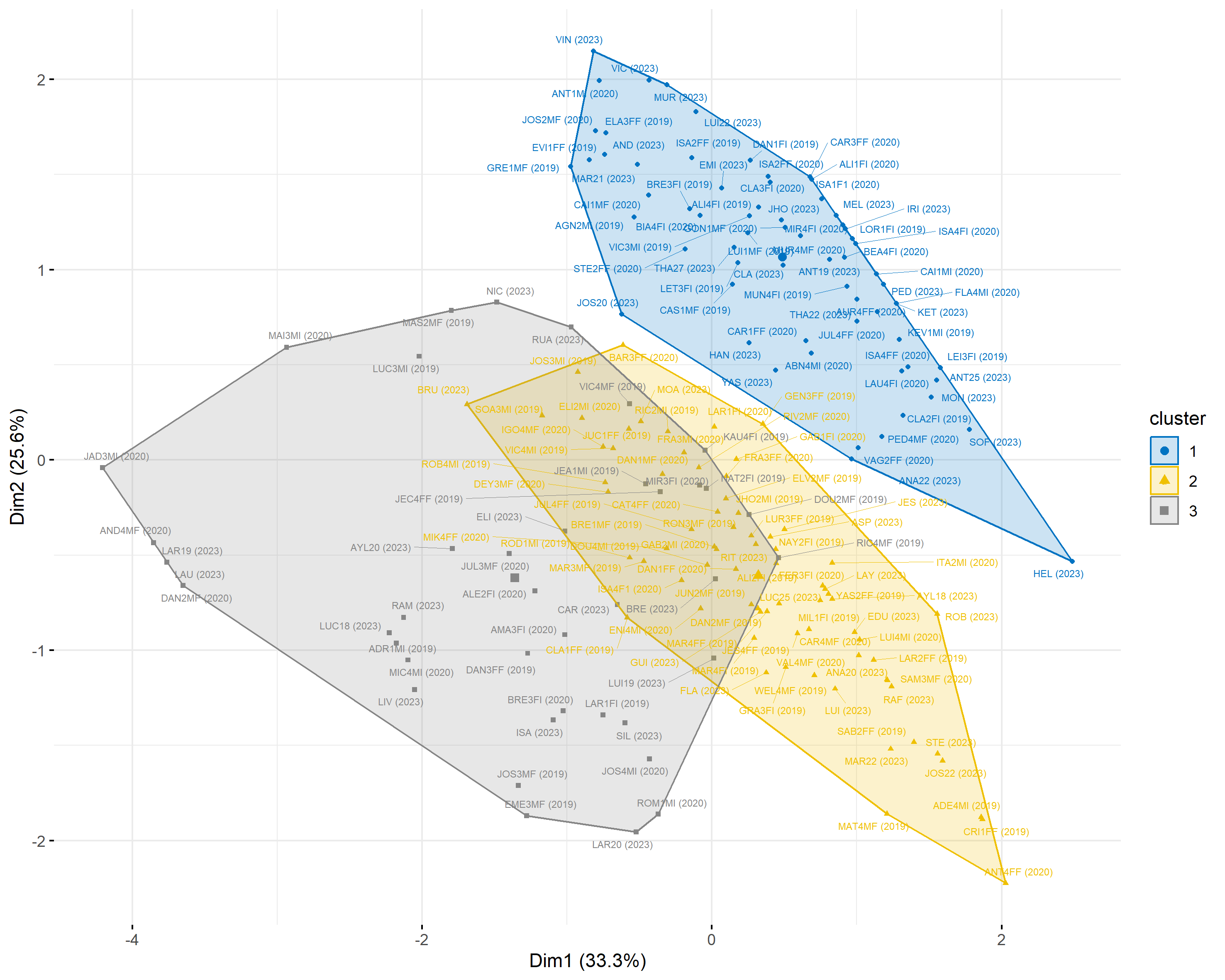}
    \caption{Clustering results}
    \label{fig:Fig3}
\end{figure*}

\subsection{Clustering}

Cluster analysis tested the hypothesis that grouping speakers based on their linguistic behavior would enable identification of their dialectal region. The hypothesis predicted that > 75\% of speakers with the same mobility profile would cluster together. Data were scaled using R's scale function, and the fviz\_nbclust function from the factoextra package indicated three clusters as the optimal number.

Principal component analysis (PCA) showed that Dim1 captured 33.3\% of variability and Dim2 captured 25.6\%, totaling 58.9\%. Results did not confirm the initial hypothesis, as most speakers with the same mobility profile did not cluster together, with the possible exception of Bahia speakers (66.6\% in Cluster 1) (Figure~\ref{fig:Fig3}).

\begin{figure*}[h]
    \centering
    \includegraphics[width=0.8\textwidth]{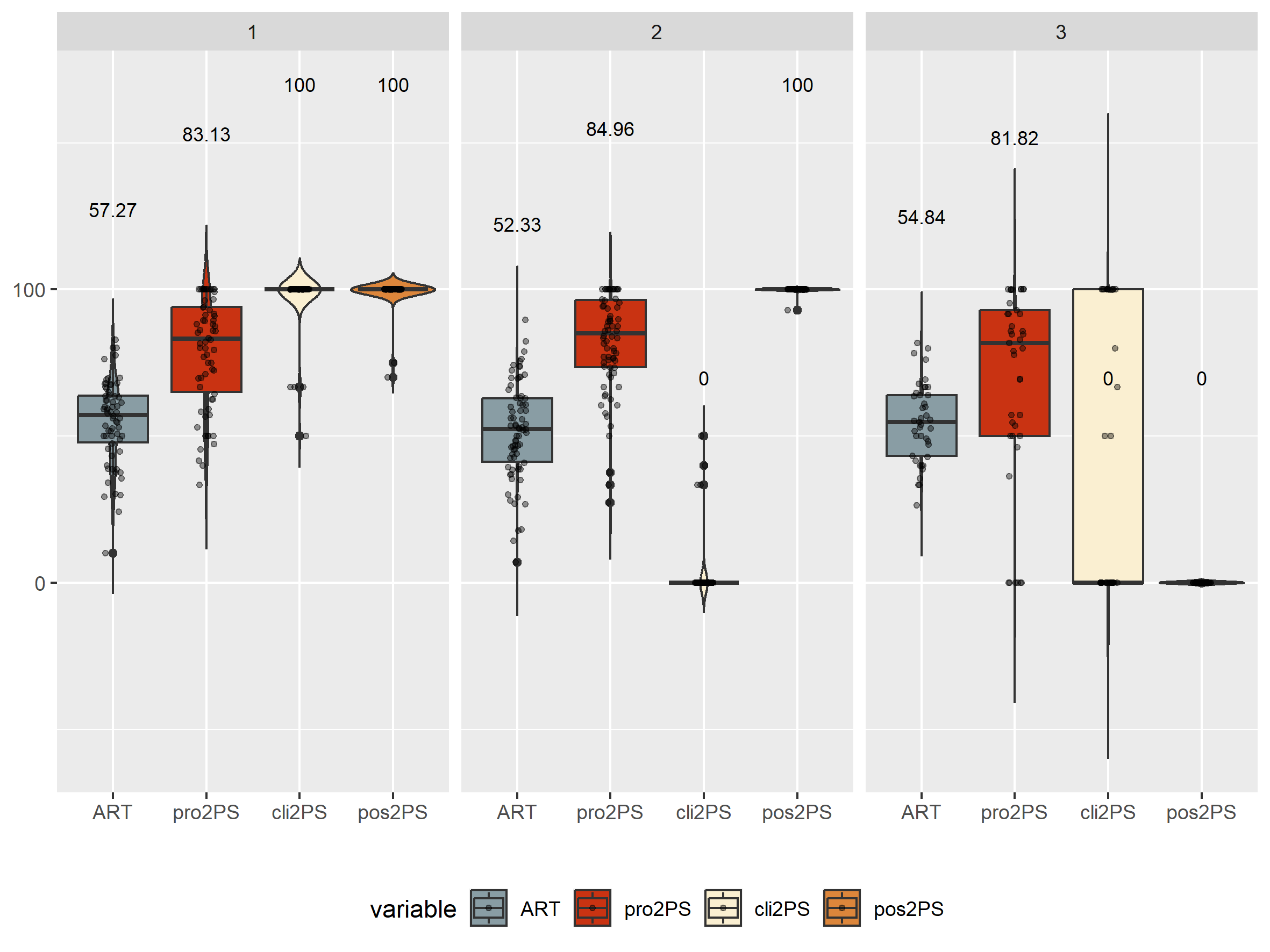}
    \caption{Mean by clusters}
    \label{fig:Fig4}
\end{figure*}

Analysis revealed three main groups (Figure~\ref{fig:Fig4}). 
Group 2 (n = 74) was characterized by lower rates of determiner absence before possessives (\textbf{det-poss} = 52.3\%), higher use of \textit{você} (\textbf{pro2P} = 85\%), low use of \textit{te} (\textbf{clit3P} = 0\%), and high use of \textit{seu} (\textbf{poss2P} = 100\%), showing heterogeneous distribution across displacement profiles. 
Group 1 (n = 66) exhibited higher determiner absence (\textbf{det-poss} = 57.3\%), high use of \textit{você} (\textbf{pro2P} = 83.1\%), \textit{te} (\textbf{clit2P} = 100\%), and \textit{seu} (\textbf{poss2P} = 100\%), with notably high numbers of speakers from Displacement 1 and Bahia. 
Group 3 (n = 41) was characterized by lower use of \textit{você} (\textbf{pro2P} = 81.8\%) and absence of \textit{te} and \textit{seu} (\textbf{poss2P} = 0\% e \textbf{clit2P} = 0\%), with no speakers from Bahia.

\section{Discussion and Conclusion}

This study investigated how morphosyntactic features co-occur within speech communities in Brazilian Portuguese, integrating perspectives from variationist sociolinguistics and NLP. By examining four 2P morphosyntactic variables — \textbf{pro2PS}, \textbf{clit2P}, \textbf{poss2PS}, and \textbf{det-poss} — in a dataset of 181 speakers and over one million words from the \textit{Falares Sergipanos} dataset, the study assessed whether patterns of covariation could reveal dialectal profiles.

Three of the four variables — \textbf{det-poss} , \textbf{pro2PS} excluding \textit{tu}), and \textbf{clit2P} — showed evidence of dialectal differentiation. Correlation analysis revealed a single significant relationship: a negative correlation between determiner absence in \textbf{det-poss} and the use of \textit{você} in \textbf{pro2PS} ($\rho = -0.18$, $p = 0.017$). This suggests that speakers who frequently omit determiners before possessives tend not to use \textit{você}, possibly reflecting regionally grounded linguistic preferences. 

From an NLP perspective, the use of morphosyntactic embeddings indicates that computational models may be able to group Brazilian Portuguese speakers according to regional patterns, even without reliance on explicit lexical markers. Cluster analysis identified three speaker groups, with the first two principal components accounting for 58.9\% of the total variance (Dim1: 33.3\%; Dim2: 25.6\%). The modeling captured structural similarities among Northeastern varieties. However, these clusters did not align categorically with mobility profiles, indicating that dialectal identity operates along subtle gradients rather than discrete boundaries. An exception was observed for speakers from Bahia, 66.6\% of whom clustered together, suggesting a degree of regional coherence.

The clustering results show that dialectal varieties reflect gradients of sociolinguistic influence, a fact that AI systems must take into account to avoid linguistic erasure and bias. These findings therefore underscore the ethical imperative of developing computational models of dialectal variation that are grounded in empirical sociolinguistic data, ensuring that language technologies recognize and respect regional varieties rather than penalizing speakers based on their linguistic profiles.

The moderate explanatory power of morphosyntactic features alone suggests that future research should incorporate additional linguistic levels, such as phonological or lexical information - as well as explore other morphosyntactic properties that may further capture patterns of dialectal variation -, or explore alternative machine-learning architectures. Furthermore, the marginal effects of macro-social variables such as gender and age indicate that regional factors constitute the primary axis of variation in this dataset. This finding aligns with previous work showing that morphosyntactic variation in Brazilian Portuguese is sensitive to geographic differentiation, especially in contexts characterized by high speaker mobility \cite{scherre2015, scherre2021, oushiro2016social, freitag2022mobility, siqueira2023}.

\section*{Limitations}

The interdisciplinary integration of sociolinguistics and NLP reveals a methodological paradox centered on sample size. From a sociolinguistic perspective, the 181 sociolinguistic interviews comprising over one million words represent a substantial dataset. 

Traditional variationist studies typically analyze dozens of speakers, making this sample exceptionally large by sociolinguistic standards. This scale enables robust statistical analyses of individual variation, social stratification, and covariation patterns across multiple morphosyntactic phenomena. The dataset provides sufficient data to examine subtle gradients of linguistic behavior and to identify dialectal distinctions that might be obscured in smaller datasets. 

From an NLP perspective, however, a dataset with 181 individuals constitute a very small dataset. Contemporary machine learning models typically require far larger datasets to avoid overfitting and to capture the full range of morphosyntactic variability. This limitation likely contributes to the moderate explanatory power of the principal components (58.9\% of variance), suggesting that additional dimensions of variation may remain undetected.

Beyond the quantitative mismatch, the qualitative provenance of speech data deserves particular attention in NLP research. Datasets constituted following sociolinguistic methodology offer irreplaceable advantages: they are collected with informed consent, comply with data protection frameworks such as the Brazilian General Data Protection Law (Lei Geral de Proteção de Dados – LGPD, Law No. 13,709/2018), and are approved by institutional ethics committees. Moreover, they are designed and curated by linguist specialists who ensure not only legal compliance but also technical rigor in speaker selection, interview protocols, transcription conventions, and metadata annotation. These dimensions – ethical, legal, and scientific – cannot be replicated by simply harvesting large volumes of speech data from the web or social media platforms, practices that are increasingly common in NLP pipelines but that raise serious concerns regarding consent, representativeness, and bias \cite{ dodge2021documenting}.

A particularly consequential limitation arises when speech data are transcribed automatically using AI-powered tools, such as Whisper \cite{radford2023robust}, without subsequent human revision by trained linguists. Although automatic speech recognition (ASR) tools offer scalability advantages, they are systematically trained on standardized, prestige varieties of language, which leads to a well-documented bias against non-standard, regional, and socially marked speech \cite{bera2024bias}. In the context of dialectal variation, this bias is not merely a performance issue: it constitutes an epistemic distortion. Dialectal features – including vowel reduction, consonant deletion, variable agreement, and prosodic patterns – are precisely the markers that are most vulnerable to being smoothed out, normalized, or outright erased during automatic transcription \cite{borges_gois_cardoso_falares}. When downstream NLP analyses, including dialectal identification and classification tasks such as those undertaken in this study, are built on automatically transcribed data without dialectal-aware correction, they risk producing results that are systematically skewed toward standard variety features, rendering invisible the very variation they seek to capture. Sociolinguistically curated corpora, with human-validated transcriptions following established conventions, are therefore not merely preferable but methodologically necessary for reliable dialectometric analyses.

In this regard, the current study also points to a broader infrastructural gap in Brazilian NLP. The relative scarcity of large-scale, openly available, and ethically constituted speech corpora of Brazilian Portuguese, particularly those representing regional, socially stratified, and underrepresented varieties, remains one of the central bottlenecks for the development of linguistically robust language technologies. The \textit{Plataforma da Diversidade Linguística Brasileira} (Brazilian Linguistic Diversity Platform) \cite{freitag2025plataforma} presents a significant step toward addressing this gap. The platform is conceived as a curated, interoperable, and ethically governed national repository of Brazilian linguistic data, designed to aggregate sociolinguistic corpora produced under rigorous methodological and ethical standards. By bringing together data from multiple research groups, spoken varieties, and regional dialects under a unified infrastructure compliant with the LGPD and guided by established principles of open science and FAIR data (Findable, Accessible, Interoperable, Reusable), the platform has the potential to provide NLP research with unprecedented access to high-quality, linguistically annotated data. Future NLP studies focused on dialectal identification, sociolinguistic profiling, and Brazilian Portuguese language modeling stand to benefit substantially from the integration of resources made available through this platform \cite{freitag2022sociolinguistic, sousa_freitag_2024, freitag2025plataforma}.

Additional limitations stem from the exclusive focus on morphosyntactic features. Although effective in capturing regional patterns without lexical cues, dialectal variation operates across multiple linguistic levels. Integrating phonological, lexical, discursive, and prosodic dimensions could improve classification accuracy but would significantly increase the annotation burden.

The restriction to university students aged 18-30 limits generalizability across age groups and social class. Moreover, the use of mobility profiles as a proxy for dialectal origin, while innovative, only partially captures the complex social and biographical factors shaping speakers’ linguistic repertoires.

Finally, the clustering analysis failed to reach the expected $\geq 75\%$ homogeneity within mobility-based groups, indicating either limits of morphosyntactic information for dialect classification or the need for methodological refinement. Alternative supervised or neural approaches may yield stronger results but require substantially larger datasets, and \textit{Plataforma da Diversidade Linguística Brasileira} could be a source.

Future research should emphasize collaborative corpus pooling, semi-automated annotation, and the application of transfer and few-shot learning techniques to reconcile sociolinguistic depth with NLP scalability \cite{freitag2022sociolinguistic, sousa_freitag_2024}.

\section*{Acknowledgments}

This research was supported by a doctoral scholarship \cite{silva_2025_covariacao} from Foundation for Support of Research and Technological Innovation of the State of Sergipe (Fapitec), and by Research Productivity Fellowship from National Council for Scientific and Technological Development (CNPq). 

\bibliography{custom}

\end{document}